\documentclass[runningheads]{llncs}
\usepackage[T1]{fontenc}
\usepackage{graphicx,verbatim}
\usepackage{booktabs}
\usepackage{multirow}
\begin{document}
\title{ProFound: A moderate-sized vision foundation model for multi-task prostate imaging}
\titlerunning{ProFound}
%
\author{Yipei Wang\inst{1,2}\orcidID{0000-0002-9589-7177} \and
Yinsong Xu\inst{1,3}\orcidID{0000-0002-1050-0143} \and
Weixi Yi\inst{1,2}\orcidID{0009-0002-9889-9853} \and
Shaheer Ullah Saeed\inst{1,2,4,5,6}\orcidID{0000-0002-5004-0663} \and
Natasha Thorley\inst{7,8}\orcidID{0000-0001-8928-895X} \and
Alexander Ng\inst{9,12}\orcidID{0000-0001-5441-2017} \and
Yukun Zhou\inst{1,10}\orcidID{0000-0002-0840-6422} \and
Wen Yan\inst{1,2}\orcidID{0000-0002-3962-5994} \and
Dean Barratt\inst{1,2}\orcidID{0000-0003-2916-655X} \and
Shonit Punwani\inst{7,8}\orcidID{0000-0002-1014-0870} \and
Veeru Kasivisvanathan\inst{9,11,12,13}\orcidID{0000-0002-0832-382X} \and
Mark Emberton\inst{11,12}\orcidID{0000-0003-4230-0338} \and
Daniel C. Alexander\inst{1,14}\orcidID{0000-0003-2439-350X} \and
Yipeng Hu\inst{1,2}\orcidID{0000-0003-4902-0486} 
}
\authorrunning{Y. Wang et al.}
%
\institute{UCL Hawkes Institute, University College London, London, UK \\ \email{yipei.wang@ucl.ac.uk} \and
Department of Medical Physics and Biomedical
Engineering, University College London, London, UK \\ \and
School of Artificial Intelligence, Beijing University of Posts and Telecommunications, Beijing, China
\\ \and
Centre for Bioengineering, Queen Mary University of London, London, UK \\ \and 
School of Engineering and Materials Science, Queen Mary University of London, London, UK \\ \and Digital Environment Research Institute, Queen Mary University of London, London, UK
\\ \and
Centre for Medical Imaging, University College London, London, UK \\ \and 
Department of Radiology, University College London Hospital NHS Foundation Trust, London, UK \\ \and
Centre for Urology Imaging, Prostate, AI and Surgical Studies (COMPASS) Research Group, Division of Surgery and Interventional Science, University College London, London, UK \\ \and
Institute of Ophthalmology, University College London, London, UK \\ \and 
Department of Urology, University College London Hospital, London, UK \\ \and 
Division of Surgery and Interventional Science, University College London, London, UK \\ \and 
Department of Urology, Comprehensive Cancer Center, Medical University of Vienna, Vienna, Austria \\ \and
Department of Computer Science, University College London, London, UK
}


  
\maketitle              
\begin{abstract}
Many diagnostic and therapeutic clinical tasks for prostate cancer increasingly rely on multi-parametric MRI. Automating these tasks is challenging because they necessitate expert interpretations, which are difficult to scale to capitalise on modern deep learning. Although modern automated systems achieve expert-level performance in isolated tasks, their general clinical utility remains limited by the requirement of large task-specific labelled datasets.
In this paper, we present ProFound, a domain-specialised vision foundation model for volumetric prostate mpMRI. ProFound is pre-trained using several variants of self-supervised approaches on a diverse, multi-institutional collection of 5,000 patients, with a total of over 22,000 unique 3D MRI volumes (over 1,800,000 2D image slices). 
We conducted a systematic evaluation of ProFound across a broad spectrum of $11$ downstream clinical tasks on over 3,000 independent patients, including prostate cancer detection, Gleason grading, lesion localisation, gland volume estimation, zonal and surrounding structure segmentation. Experimental results demonstrate that finetuned ProFound consistently outperforms or remains competitive with state-of-the-art specialised models and existing medical vision foundation models trained/finetuned on the same data. 
In this paper, we share real-world experience in developing and assessing ProFound, with a set of rigorous, large-scale clinical evaluations.
To facilitate deployment and finetuning under heterogeneous clinical compute constraints, we adopt a moderate-sized, efficient architecture. ProFound offers a scalable framework for advancing compute-and-data efficiency in prostate oncology. All model weights, implementation 
as well as an interactive interface are publicly available at \url{https://github.com/pipiwang/ProFound} and \\ \url{https://huggingface.co/spaces/Anonymise/ProFound}.

\keywords{Prostate Cancer \and mpMRI \and Foundation model.}

\end{abstract}
\section{Introduction}
Prostate cancer remains one of the leading causes of cancer-related mortality in men, and early, accurate diagnosis is crucial for improving patient outcomes.
In current clinical pathways, 
multi-parametric MRI (mpMRI) is increasingly used across multiple stages of care, including pre-biopsy risk stratification, lesion targeting, and surgical and radiotherapy planning. Interpretation of mpMRI and the associated downstream decisions remain highly dependent on expert radiological assessment, which can introduce variability in reporting. In this context, deep learning for prostate MRI has shown progress in several clinical tasks, including cancer detection/classification and lesion or anatomy segmentation \cite{xu2024ordinal,yan2025semi,sanford2020deep}. However, automation using deep learning often requires a large volume of labelled data, which becomes increasingly infeasible for wider adoption in different centres and for diverging tasks. 

General-purpose radiology foundation models aim to reduce the data requirements for specific downstream tasks by providing an easy-to-generalise model trained across many organs and modalities \cite{wu2025radfm,zheng2024raddiag} and have shown potential in achieving better performance for individual applications. However, achieving strong performance often still requires substantially larger amounts of pretraining data, even millions \cite{wu2025radfm}. 
In addition, for tasks involving prostate MR, model performance remains insufficient for clinical use due to the under-representation of prostate mpMRI, likely because prostate mpMRI is a highly specialised multi-sequence protocol and its widespread standardisation in clinical practice (e.g., through PI-RADS \cite{vargas2016updated}) is relatively new.

Recent efforts have begun to explore domain-specialised foundation models for prostate imaging. ProViCNet \cite{lee2025prostate} introduced a prostate-specific vision transformer for clinically significant cancer detection on mpMRI and TRUS, leveraging patch-level contrastive learning to enhance lesion discrimination and demonstrating strong detection performance.
ProstNFound \cite{wilson2024prostnfound} adapted medical foundation models to micro-ultrasound by incorporating domain-specific texture prompts and clinical metadata, achieving competitive ultrasound-based cancer detection performance and highlighting the importance of injecting modality-specific priors into foundation models. 
Despite these advances, existing approaches remain largely task-centric and/or modality-specific, focusing on a diagnostic objective (primarily cancer detection) rather than learning a unified prostate representation that can transfer across the broader set of clinical tasks required in prostate cancer management.
In practice, beyond cancer detection, mpMRI has an arguably wider range of clinical applications, including lesion delineation for therapy planning, localisation at clinically actionable region levels, grading/risk stratification, etc. A prostate-dedicated, volumetric foundation model trained directly on large-scale prostate mpMRI and evaluated systematically across these diverse downstream tasks remains unexplored. 

To examine whether a practically feasible pretraining data scale can still support broad downstream utility, we present ProFound, a vision foundation model pre-trained using self-supervised masked autoencoding \cite{he2022masked} on a moderate-scale, multi-institutional collection of prostate mpMRI data drawn from diverse patient populations and clinical cohorts (e.g., screening, diagnostic, and treatment-planning pathways) and using T2w, DWI, and ADC sequences to learn prostate-specific volumetric representations. We evaluate ProFound across multiple downstream clinical tasks spanning classification, segmentation, localisation, and regression, and compare against (i) task-specific models trained from scratch and (ii) existing generalist medical foundation models as well as general vision pretraining baselines. Further, we study ProFound’s efficiency under limited available finetuning data and limited finetuning regimes, and report systematic clinical evaluation including comparisons to radiologist performance and histopathology reference standard. Integrating deep learning into real-world clinical workflows requires models that can run efficiently and securely on standard hospital hardware without prohibitive computational costs. These have motivated our moderately-sized, efficient, and open-source ProFound, for a scalable framework for compute- and data-efficient prostate imaging analysis. ProFound is open-sourced at \url{https://github.com/pipiwang/ProFound}.

\begin{figure}[htbp]
    \centering
    \includegraphics[width=\textwidth]{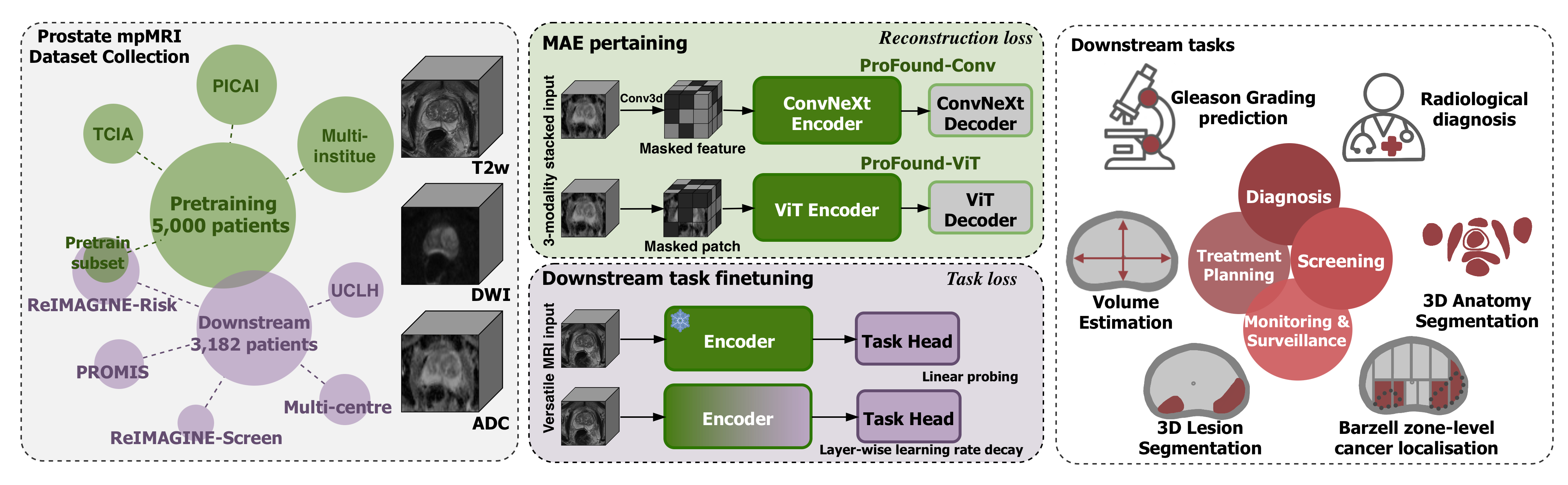}
    \caption{Overview of the proposed prostate mpMRI foundation model.
    }
    \label{fig:overview}
\end{figure}

\section{Method}\label{sec:method}
\subsection{Data}
\noindent \textbf{Pretrain dataset}
ProFound is pretrained on a large and diverse collection of prostate MRIs aggregated from multiple datasets, including two public datasets: the PICAI challenge dataset \cite{saha2024artificial} and the Cancer Imaging Archive (TCIA) dataset \cite{Choyke_Turkbey_Merino_Wood_2025}, comprising $1,500$ and $863$ patients respectively; and two private datasets: the ReIMAGINE Risk study dataset \cite{marsden2021update} with $491$ patients and a multi-institute dataset \cite{min2025segmentation} with $2,146$ patients. Three MRI modalities were used during pretraining, including T2-weighted (T2w), high b-value diffusion-weighted (DWI) and apparent diffusion coefficient (ADC) MRI. The number of available mpMRI scans varied across patients, ranging from 1 to 28 three-modality triplets per patient, resulting in a total of $8,893$ pretraining triplets from $5,000$ patients. 
All images were resampled to an isotropic spatial resolution of $1mm \times 1mm \times 1mm$, spatially aligned across modalities, and intensity-normalised to a consistent range. We applied data augmentation consisting of random isotropic zooming, 
random spatial cropping, and random flipping along the axial axis.

\noindent \textbf{Downstream task finetuning and evaluation dataset}
Various downstream tasks were performed on the following prostate mpMRI datasets: the multi-institute ReIMAGINE Risk from $1,028$ patients \cite{marsden2021update}, the ReIMAGINE Screen dataset from $304$ patients \cite{Marsdene2021screen}, the PROMIS study dataset from $566$ patients \cite{wang2025promis}, the UCLH dataset from $695$ patients \cite{yan2024combiner}, and a multi-centre pelvis MR dataset with $589$ patients \cite{li2023fewshot}. All datasets were divided into finetuning/training, validation and test sets in a 7:1:2 ratio. The downstream task datasets follow the same preprocessing pipeline as the pretraining dataset.

\begin{table}[tb]
\centering
\caption{Description of ProFound downstream tasks and extended analysis. Cat. = Category, Cls = classification, Seg = segmentation, Reg = regression, Loc = localisation, mul = multi-class, bin = binary, GG = Gleason group, FM = Foundation model  
}
\label{tab:task}
\resizebox{\textwidth}{!}{%
\begin{tabular}{lllll}
\toprule
Task name & Description & Dataset & Subtask & Cat. \\ \midrule
PIRADS-Cls & Automating radiologist task by predicting  & ReIMAGINE & PIRADS$\leq3$, 3, 4, 5 & mul-Cls \\
 &  PIRADS score & -Risk & PIRADS$<3$ vs. $\geq3$ & bin-Cls \\
 &  &  & PIRADS$\leq3$ vs. $>3$ & bin-Cls \\ 
Grading-Cls & Predicting Gleason Grade Group on patients  & PROMIS & GG$\leq6$,3+4,4+3,8,\textgreater{}8 & mul-Cls \\
 &  with MR-positive lesions &  & GG$\leq3+4$ vs. $\leq6$ & bin-Cls \\
 &  &  & GG$\leq4+3$ vs. $\leq3+4$ & bin-Cls \\
LesionSegAll & Delineating lesion contour on all patients & UCLH &  & bin-Seg \\ 
LesionSegRad & Delineating lesion contour on radiology-positives & \multicolumn{2}{l}{PROMIS (PIRADS\textgreater{}2 subset)} & bin-Seg \\
AnatomySeg & Multi-structure segmentation for focal therapy & multi-centre &  & mul-Seg \\
Volume-Reg & Prostate volume estimation & UCLH &  & Reg \\
PCa-Loc & Histopathological PCa detection on Barzell zones & PROMIS &  & Loc \\
Tang et al\cite{tang2025impact} & Impact of image quality on FM finetuning & PRIME &  &  \\
Sidiqi et al\cite{sidiqi2026continual} & Investigating task-aligned continual pretraining & \multicolumn{2}{l}{PROMIS, UCLH, ReIMAGINE-Risk} &  \\
Huang et al\cite{huang2026understanding} & Studying task alignment and transferability & \multicolumn{2}{l}{PROMIS, multi-centre} & \\
Wang et al\cite{wang2026fine} & Interpreting downstream task finetuning & \multicolumn{2}{l}{UCLH, ReIMAGINE-Risk}  &  \\ \bottomrule
\end{tabular}%
}
\end{table}
\subsection{Pretraining}
In this work, we adopt the masked autoencoder (MAE) \cite{he2022masked} framework to pretrain large-scale volumetric foundation models using two variants under a unified self-supervised objective. Given a combination of T2w, high-b DWI, and ADC from one patient, the 3D mpMRI sequences are stacked as channels to form a 3-modality input. A high proportion of patches or image volumes are masked to train the network to reconstruct the missing content from the remaining visible context. 
We investigate two 3D masked autoencoder architectures: 1) \textbf{ProFound-ViT} is a ViT-based 3D MAE adapted from ViT-B  \cite{dosovitskiy2020image}. It tokenises the 3D volumes into non-overlapping cubic patches, processes visible patches with Transformer blocks using fixed 3D sine–cosine positional embeddings, and reconstructs masked patches with a lightweight decoder following the standard asymmetric MAE design. The reconstruction loss is computed only over the masked image volumes;
2) \textbf{ProFound-Conv} is a fully convolutional 3D MAE based on ConvNeXtV2 \cite{woo2023convnext}. Rather than tokenising patches, it preserves a dense 3D feature grid and applies masking directly in the spatial domain. 
The encoder uses sparse convolutions to operate primarily on visible voxels for improved efficiency, while a lightweight convolutional decoder restores masked regions using learned mask tokens. The reconstruction loss is applied only to masked regions.

\subsection{Downstream task finetuning}
A wide range of downstream tasks were chosen as described in Tab~\ref{tab:task}. 
For classification, we use two fully connected layers with an intermediate batch normalisation layer. For segmentation, we assess two commonly used architectures, UperNet \cite{xiao2018unified} and UNetR3D \cite{hatamizadeh2022unetr}. For AnatomySeg, ProFound is adapted to take only T2w by zeroing out the other two input channels. The localisation task predicts cancer presence on Barzell zones of the prostate \cite{wu2025ai}, which follows the same design as the segmentation task (UNetR3D), with an additional layer that performs zone-wise feature aggregation and subsequent classification. For regression, we adopt a batch normalisation layer followed by a fully connected layer.

\subsection{Experiments for evaluating downstream tasks}
To assess the impact of prostate-specific pretraining and data efficiency, we evaluate ProFound against three categories of models and various adaptation regimes:

\noindent\textbf{Foundation Model Baselines}: We compare against RadFM \cite{wu2025radfm} (pretrained on 16M radiological pairs), RadDiag \cite{zheng2024raddiag} (pretrained on 192k multi-modal scans), and a DINOv2 \cite{oquab2024dinov} baseline trained on our pretrain dataset to isolate the effects of the MAE objective and architecture.

\noindent\textbf{Specialised Models}: To quantify pretraining advantages, ProFound backbones (ViT and ConvNeXtV2) are compared against identical architectures trained from scratch and standard 3D medical imaging baselines: ResNet18 (Cls/Reg tasks) and UNet (Seg/Loc tasks) using the same data for finetuning.

\noindent\textbf{Adaptation Strategies}: We evaluate two regimes: linear probing (frozen encoder) and full finetuning using layer-wise learning rate decay (LLRD) for stable feature preservation.

\noindent\textbf{Data Efficiency}: Models are tested under constrained conditions (8 to 128 samples) to simulate clinical scenarios where annotations are scarce.

\noindent\textbf{Evaluation Metrics}:
Performance is averaged across three runs. For ordinal multi-class tasks, we use Quadratic Weighted Kappa and binary AUC (with sensitivity/specificity at 80\% fixed operating points). Segmentation is evaluated via the Dice score, and volume estimation via Mean Absolute Error (MAE). Statistical significance is determined using paired t-tests.



\noindent\textbf{Implementation}:
The mask ratios for ProFound-ViT and ProFound-Conv during pretraining were set to $0.75$ and $0.6$, respectively, following \cite{he2022masked} and \cite{woo2023convnext}.
The best performing model is selected based on the validation set evaluation within $100$ epochs of finetuning or training. The classification, segmentation, and regression tasks are trained using cross-entropy loss, Dice with cross-entropy loss, and Mean Squared Error loss respectively. Note that in less data settings, UNet is unstable when trained with Dice with cross-entropy loss, therefore all comparison models are trained with Dice loss. 
ProFound-ViT and ProFound-Conv are pretrained on 4 RTX8000 GPUs for 10 days and 4 A6000 GPUs for 14 days respectively.
Further implementation details including pretrain/finetune epochs, learning rates, etc., can be found in the open-source repository.

\section{Results}

\noindent\textbf{ProFound achieves SOTA performance in segmentation tasks:}
For prostate lesion segmentation on all patients with MRI scans, ProFound-Conv achieves the best performance with a Dice of ($0.429\pm0.009$), significantly outperforming the strongest specialised baseline UNet ($0.416\pm0.004$, $p=0.001$). Overall, models leveraging foundation pretraining tend to outperform their train-from-scratch counterparts, highlighting the value of large-scale pretraining for prostate MRI representation learning. Notably, ProFound-Conv also significantly exceeds DINOv2 ($0.415\pm0.007$, $p<0.0001$), despite DINOv2 being pretrained on the same dataset as ProFound.  

For lesion segmentation on radiologist-positive patients, Dice scores decrease across all methods. This could partially be attributed to the smaller dataset size and that the UCLH dataset is specifically collected for patients enrolled in focal therapy, which are expected to include larger and more conspicuous lesions, while the PROMIS dataset is collected on a biopsy-blind diagnostic cohort. Nonetheless, ProFound-Conv with frozen encoder ranks highest(0.297$\pm$0.005), closely followed by other ProFound variants $(0.296\pm0.009$ and $0.293\pm0.008)$. 

For the prostate anatomy segmentation task, ProFound-ViT with fixed encoder achieves the top Dice of $0.931\pm0.001$, significantly outperforming RadDiag ($0.921\pm0.001$).
Across segmentation tasks, ProFound models with UNetR3D head generally outperform those using UperNet, with all $p < 0.0001$ for LesionSegAll and AnatomySeg tasks (except for ProFound-ViT-UNetR3D on AnatomySeg). 

\begin{figure}[tbp]
    \centering
    \includegraphics[width=\textwidth]{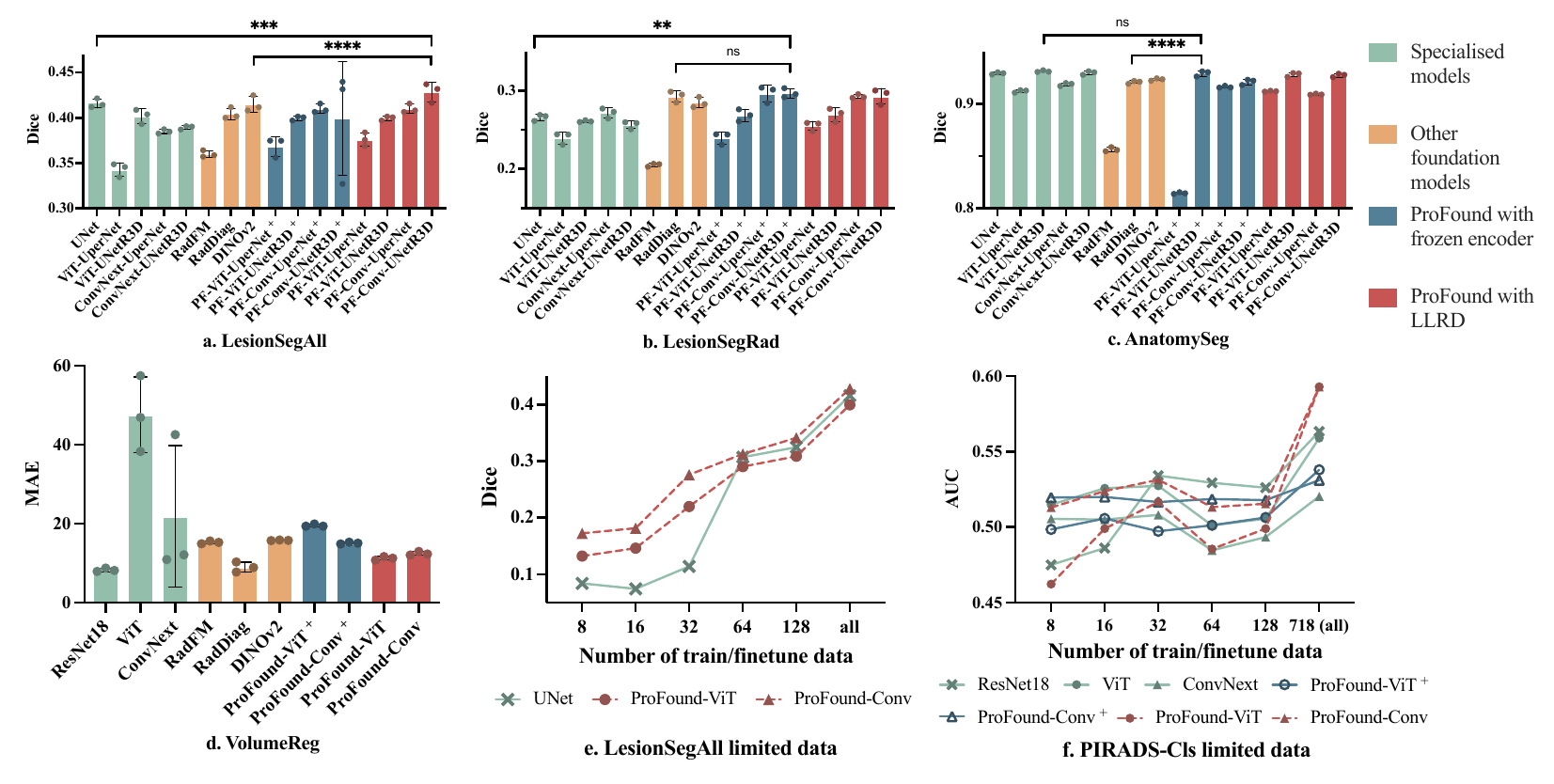}
    \caption{Performance on representative downstream tasks. ProFound achieves the highest Dice scores on both lesion segmentation tasks, while remaining comparable to SOTA specialised models on anatomy segmentation and prostate volume regression. ProFound shows a clear advantage in limited-label settings, demonstrating strong data efficiency. We also include a challenging task where ProFound provides more nuanced gains.
    }
    \label{fig:seg}
\end{figure}

ProFound demonstrates strong data efficiency. In limited-data experiment, ProFound achieves a Dice score of $0.276$ with only 32 finetuning cases, and consistently outperforms UNet across all low-data settings, as shown in Fig~\ref{fig:seg}.

\noindent\textbf{ProFound supports prostate cancer detection:}
For PIRADS score prediction, ProFound-Conv achieved the strongest ordinal agreement with the ground truth (QWK $= 0.285$), outperforming DINOv2 ($0.227$), ViT ($0.224$), and ResNet18 ($0.207$). Several baselines (e.g., ConvNeXt: $0.010$; RadFM :$-0.002$) demonstrate limited ordinal consistency despite reasonable binary discrimination performance. For binary classification of radiologist-visible cancer (PIRADS $\geq 3$ vs.\ $<3$), ProFound-Conv attains the highest AUC (76.1\%), followed by ProFound-ViT (75.6\%) and RadFM (73.4\%). At clinically meaningful operating points, RadFM yields the highest sensitivity at 80\% specificity (62.6\%), while ProFound-ViT and RadDiag achieve the highest specificity at 80\% sensitivity (55.6\%). 

For Gleason Grading classification among MR-positive patients, ProFound-Conv again provides the best ordinal agreement (QWK $=0.169$), followed by RadDiag ($0.147$) and DINOv2 ($0.100$). In contrast, multiple supervised baselines exhibited near-zero or negative QWK (e.g., ResNet18: $-0.002$; ViT: $-0.035$), indicating poor alignment with grading labels. For binary classification of Gleason Grade Group $\geq3$, ProFound-Conv achieves the highest AUC (73.5\%), surpassing RadDiag (72.0\%). At fixed operating points, ProFound-Conv achieves the highest specificity at 80\% sensitivity (60.4\%), whereas RadDiag yields the highest sensitivity at 80\% specificity (50.4\%).
As summarised in Table~\ref{tab:multicls}, overall, ProFound-Conv offers the most balanced performance across ordinal agreement and binary discrimination.

For Barzell zone-level lesion detection, ProFound-Conv excel in all evaluation metrics, with all ProFound variants improving over both foundation models and supervised models, except for RadDiag which achieves comparable performance. Notably, under the same clinical definition, radiologist readings achieve 39.8\% specificity at 80\% sensitivity, which is lower than all ProFound variants.


\noindent\textbf{Other results on model interpretability, continual pretraining and further downstream tasks:}
For prostate volume estimation tasks, ResNet18 achieves the lowest MAE (ml) of 8.28, followed by RadDiag (8.00). ProFound variants show comparable performance of 11.25 and 12.44.

\begin{table}[hbt]
\centering
\caption{Performance comparison for PIRADS score classification, Gleason grade group classification, and localised cancer detection task. Spec = Specificity at 80\% sensitivity; Sens = Sensitivity at 80\% specificity, $^{+}$ denotes ProFound with frozen encoder (linear probing). The baseline model is ResNet18 for PIRADS-Cls and Grading-Cls tasks, and UNet for the PCa-Loc localisation task. The unit for the number of updated parameters is million.}
\label{tab:multicls}
\resizebox{\textwidth}{!}{%
\begin{tabular}{llllllllllcl|llllc} \toprule
 & \multicolumn{4}{l}{PIRADS-Cls} &  & \multicolumn{4}{l}{Grading-Cls} & Cls &  &  & \multicolumn{3}{l}{PCa-Loc} & Loc \\ \cline{2-5} \cline{7-10} \cline{14-16}
 & QWK & AUC & Spec & Sens & & QWK & AUC & Spec & Sens & Params &  &  & AUC & Spec & Sens & Params \\ \midrule
Baseline & 0.207 & 73.2 & 50.0 & 49.1 &  & -0.002 & 59.1 & 35.4 & 22.4 & 33.3 &  &  & 61.3 & 41.4 & 25.0 & 4.8 \\
ViT & 0.224 & 72.1 & 36.1 & 56.1 &  & -0.035 & 47.4 & 20.8 & 13.2 & 123.0 &  &  & 60.5 & 41.3 & 26.2 & 136.3 \\
ConvNeXt & 0.010 & 59.7 & 38.9 & 22.2 &  & 0.087 & 59.6 & 39.6 & 21.9 & 29.5 &  &  & 62.4 & 43.2 & 30.0 & 39.5 \\
RadFM & -0.002 & 73.4 & 47.2 & \textbf{62.6} &  & 0.013 & 55.7 & 25.0 & 18.9 & 66.4 &  &  & 60.3 & 38.5 & 22.7 & 79.6 \\
RadDiag & 0.076 & 73.3 & \textbf{55.6} & 54.4 &  & 0.147 & 72.0 & 47.9 & \textbf{50.4} & 33.3 &  &  & 63.5 & 42.7 & 27.0 & 119.4 \\
DINOv2 & 0.227 & 72.9 & 44.4 & 50.9 &  & 0.100 & 60.7 & 41.7 & 25.0 & 92.4 &  &  & 62.1 & 41.2 & 28.4 & 112.7 \\
ProFound-ViT$^{+}$ & 0.246 & 67.4 & 44.4 & 43.3 &  & 0.035 & 48.1 & 31.3 & 13.6 & 0.2 &  &  & 63.1 & 45.4 & 31.4 & 13.5 \\
ProFound-Conv$^{+}$ & 0.191 & 50.7 & 30.6 & 18.7 &  & 0.034 & 50.6 & 29.2 & 15.8 & 0.2 &  &  & 62.2 & 43.3 & 28.7 & 10.2 \\
ProFound-ViT & 0.047 & 75.6 & \textbf{55.6} & 58.5 &  & -0.063 & 49.8 & 25.0 & 16.2 & 123.0 &  &  & 63.1 & 45.4 & 31.4 & 136.3 \\
ProFound-Conv & \textbf{0.285} & \textbf{76.1} & 47.2 & 60.2 &  & \textbf{0.169} & \textbf{73.5} & \textbf{60.4} & 49.6 & 29.5 &  &  & \textbf{66.0} & \textbf{46.5} & \textbf{33.4} & 39.5 \\ \bottomrule
\end{tabular}%
}
\end{table}
The ProFound model was open-sourced prior to writing this report, other studies reported the impact of image quality on ProFound finetuning \cite{tang2025impact}, correlation between pretraining-downstream task alignment and foundation model transferability \cite{huang2026understanding}, the effect of continual pretraining on downstream task performance \cite{sidiqi2026continual}, as well as interpreting changes during downstream task finetuning \cite{wang2026fine}. These developments are not contributions of the present work, however, they provide useful context and offer complementary reference for the results reported in this paper.

\section{Discussion and Conclusion}
The segmentation results suggest that prostate-specific volumetric pretraining provides consistent gains over both specialised models trained from scratch and generalist foundation models. ProFound outperforms DINOv2 even when pretrained on the same dataset, indicating that the 3D MAE objective and architectural choices better capture prostate-relevant structure for dense prediction. We also observe that frozen-encoder adaptation can be competitive, particularly in lesion segmentation on smaller cohorts, supporting the practical utility of ProFound in limited label-and-compute finetuning regimes.

Across detection- and grading-based tasks, ProFound-Conv consistently improves ordinal agreement (QWK) while maintaining strong discrimination (AUC), suggesting that prostate-specific volumetric pretraining learns representations aligned with clinically meaningful disease ordering rather than only binary separability. The observation that some generalist models achieve reasonable AUC yet near-zero/negative QWK highlights the importance of evaluating ordinal consistency for PIRADS and Gleason-related endpoints. The improved zone-level localisation relative to radiologist performance suggests ProFound may provide complementary decision support for localised targeting and structured reporting.

A notable strength of this study is the heterogeneity of the multi-centre datasets used for pretraining and downstream evaluation. 
Cohorts differ in acquisition protocols, scanners, patient selection, and annotation type/quality, introducing realistic domain shift and label noise (e.g., inter-reader variability in lesion boundaries and PI-RADS ratings \cite{brembilla2020interreader}). 
ProFound’s consistent performance gains across multiple benchmarks suggest that the learned representations transfer effectively across diverse real-world settings rather than relying on narrowly curated, homogeneous datasets.

This study represents the first version of ProFound, with ongoing development curating larger data sets of higher quality, for pretraining and even broader validation. By fully open-sourcing our open-weight models for specialised clinical needs, this work offers a scalable, data-and compute-efficient foundation to advance the state of the art in prostate oncology.


\section{Acknowledgment}
This work was supported by the National Institute for Health Research (NIHR) University College London Hospitals (UCLH) Biomedical Research Centre (BRC). This work was also supported by the International Alliance for Cancer Early Detection, an alliance between Cancer Research UK [C28070/A30912; C73666/A31378; EDDAPA-2024/100014], Canary Center at Stanford University, the University of Cambridge, OHSU Knight Cancer Institute, University College London and the University of Manchester.

\bibliographystyle{splncs04}
\bibliography{ref}
%

\end{document}